# Prediction of Atomization Energy Using Graph Kernel and Active Learning

Yu-Hang Tang*  and Wibe A. de Jong

Lawrence Berkeley National Laboratory, Berkeley, California 94720, USA

**Abstract**   Data-driven prediction of molecular properties presents unique challenges to the design of machine learning methods concerning data structure/dimensionality, symmetry adaption, and confidence management. In this paper, we present a kernel-based pipeline that can learn and predict the atomization energy of molecules with high accuracy. The framework employs Gaussian process regression to perform predictions based on the similarity between molecules, which is computed using the marginalized graph kernel. To apply the marginalized graph kernel, a spatial adjacency rule is first applied to convert molecules into graphs whose vertices and edges are labeled by elements and interatomic distances, respectively. We then derive formulas for the efficient evaluation of the kernel. Specific functional components for the marginalized graph kernel are proposed, while the effect of the associated hyperparameters on accuracy and predictive confidence are examined. We show that the graph kernel is particularly suitable for predicting extensive properties because its convolutional structure coincides with that of the covariance formula between sums of random variables. Using an active learning procedure, we demonstrate that the proposed method can achieve a mean absolute error of $0.62 \pm 0.01$ kcal/mol using as few as 2000 training samples on the QM7 data set.

**Keywords**   *data-driven modeling, structured data, labeled graph, kernel method, quantum mechanics, computational chemistry*

## 1   Introduction

The chemistry and materials communities have embraced data science and machine learning to bring about revolutionizing solutions to long-standing challenges in molecular modeling, optimal experiment design, and high-throughput structure screening [1–5]. One particularly promising application of machine learning techniques is to train predictive models for molecular properties that are otherwise only available through expensive dynamics simulations and quantum mechanical calculations. This type of regression tasks typically entail quality requirements concerning:

1. accuracy: While the atomization energy of drug-like ligands are on or above the order of $10^3$ kcal/mol, it is the binding energy, *i.e.* the difference between the energy of two ligand-protein complexes, that is sought for applications such as drug screening. It is therefore expected that a good predictive model should carry less than 1 kcal/mol absolute error, or $10^{-3}$ relative error;

2. smoothness: The prediction must be at least twice differentiable with regard to the molecular geometry for a model to be usable as a force field for dynamics simulations;

3. outlier handling: Inevitably, a trained model could encounter outliers which triggers extrapolation far beyond the training domain. Should the predictive model fail to maintain a desired level of accuracy in this situation, a fail-safe mechanism should kick in to let the end user be aware of the potential quality degradation and take remedy actions;

4. symmetry adaptation: The predictive model should be able to recognize configurations and conformations that are merely different by a permutation and/or a rigid body transformation, which only cause covariant changes to the target property. Symmetry adaptation could take advantage of such knowledge to simultaneously improve accuracy, smoothness, and generalization.

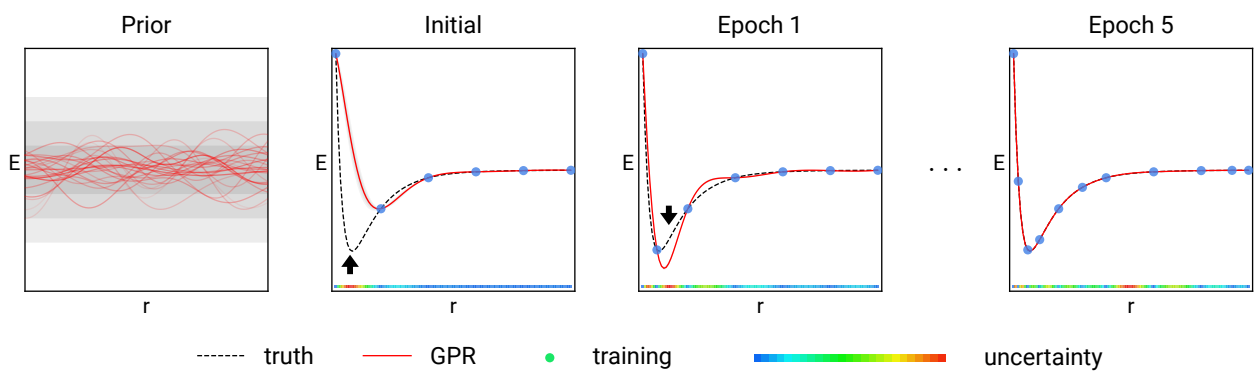

Figure 1: An active learning procedure is carried out to learn the Lennard-Jones potential between two atoms. By iteratively adding the point with the largest predictive uncertainty into the training set, the GPR model quickly converges to the ground truth using a relative small amount of samples.

Gaussian process regression (GPR) [6] is a robust regression method that features high accuracy, strong smoothness guarantee, and built-in uncertainty estimation. It is a Bayesian inference method that models the similarity among the sample points as covariance between random variables indexed by the input space. The probabilistic model generated by GPR can make not only a point estimate about an unknown sample, but also the associated posterior variance. Since the predictive variance vanishes around training samples and grows in regions that are not supported by the sample data, it can be conveniently used as a measure of predictive uncertainty for outlier detection. As shown in fig. 1, the GPR predictive uncertainty can be readily used to implement an active-learning protocol where test samples with large uncertainties are sought and added to the training set in an iterative process.

A critical component in a GPR model is a function that estimates the covariance of target values between pairs of data samples. In the context of molecular property prediction, a covariance function translates structural similarity into correlations between the target properties using the similar property principle [7]. In fact, covariance functions belong to the family of similarity kernels, which are bivariate functions that compute the closeness of two vectors in some normed vector space. A kernel is also a convenient place for encapsulating symmetry adaption mechanisms that are transparent to subsequent GPR computations. The two most well-known kernels on Euclidean spaces are probably the dot product kernel and the square exponential kernel, which defines similarity based on angle and distance, respectively.

Meanwhile, designing valid kernels on non-Euclidean spaces as those spanned by molecular configurations and conformations is an inherently challenging task, because the kernel must satisfy certain properties such as symmetry and positive-definiteness to ensure convexness of the resulting machine learning model [6]. An approach that is commonly used in existing implementations [8, 9] is to apply an Euclidean kernel to fixed-length feature vectors that are computed using a molecular fingerprinting algorithm. Examples of this approach include the atom-centered symmetry function fingerprints [10], the Coulomb matrix [11], the density-encoded canonically-aligned fingerprint [12], the encoded bonds descriptor [13] and so on. The smooth overlap of atomic positions kernel [14] used in practice also effectively computes the inner product using a fixed-length vector representation of the spherical power spectra of atomic mass density.

Despite the success of the feature vector approach, it is important to point out that squashing a molecule, which is intrinsically a non-linear structure with a variable number of degrees of freedom, into a fixed-length feature vector could be a detour that compromises the generalizability of machine learning models across the diverse chemical space. This is likely the reason why fingerprints are prevalently used to describe *local* atomistic neighborhoods that possess bounded complexities. In contrast, an intrinsically non-linear data structure, such as a graph, could be a powerful and intuitive representation to capture all the geometric and topological information of whole molecules [15, 16]. Moreover, there has been extensive work on the design of graph kernels, which compute the similarity between graphs without resorting to a fixed length intermediate presentation [17–23]. A particularly attractive feature of the graph kernels is their ability to adapt to graphs with arbitrary numbers of vertices and edges.

In this work, we present how a particular type of graph kernel, *i.e.* the marginalized graph kernel [24], can be integrated into a Gaussian process regression-based active learning pipeline for molecular atomization energy prediction. In section 2, we introduce the mathematical formulations for the techniques used in this study, including the marginalized graph kernel, Gaussian process regression, and active learning. In section 3, we present computational setup and results. We conclude the work in section 4.

## 2 Method

### 2.1 Overview

In fig. 2, we present an overview of the machine learning pipeline proposed in this work. Given a set of training molecules, a GPR model can be constructed using the pairwise similarity matrix among the molecules to fit for the assocaited target values. To compute the similarity, the molecules are converted into graphs with vertices labeled by the atoms and edges encoding interatomic distances. The marginalized graph kernel is then applied to average over the similarities of all paths generated from simultaneous random walks on each pair of graphs. Predictions for the energy of new molecules can be made using the pairwise similarity matrix among the new molecules and the cross-similarity matrix between the new molecule and the training molecules. In the subsections below, we will explain in detail each component of the procedure.

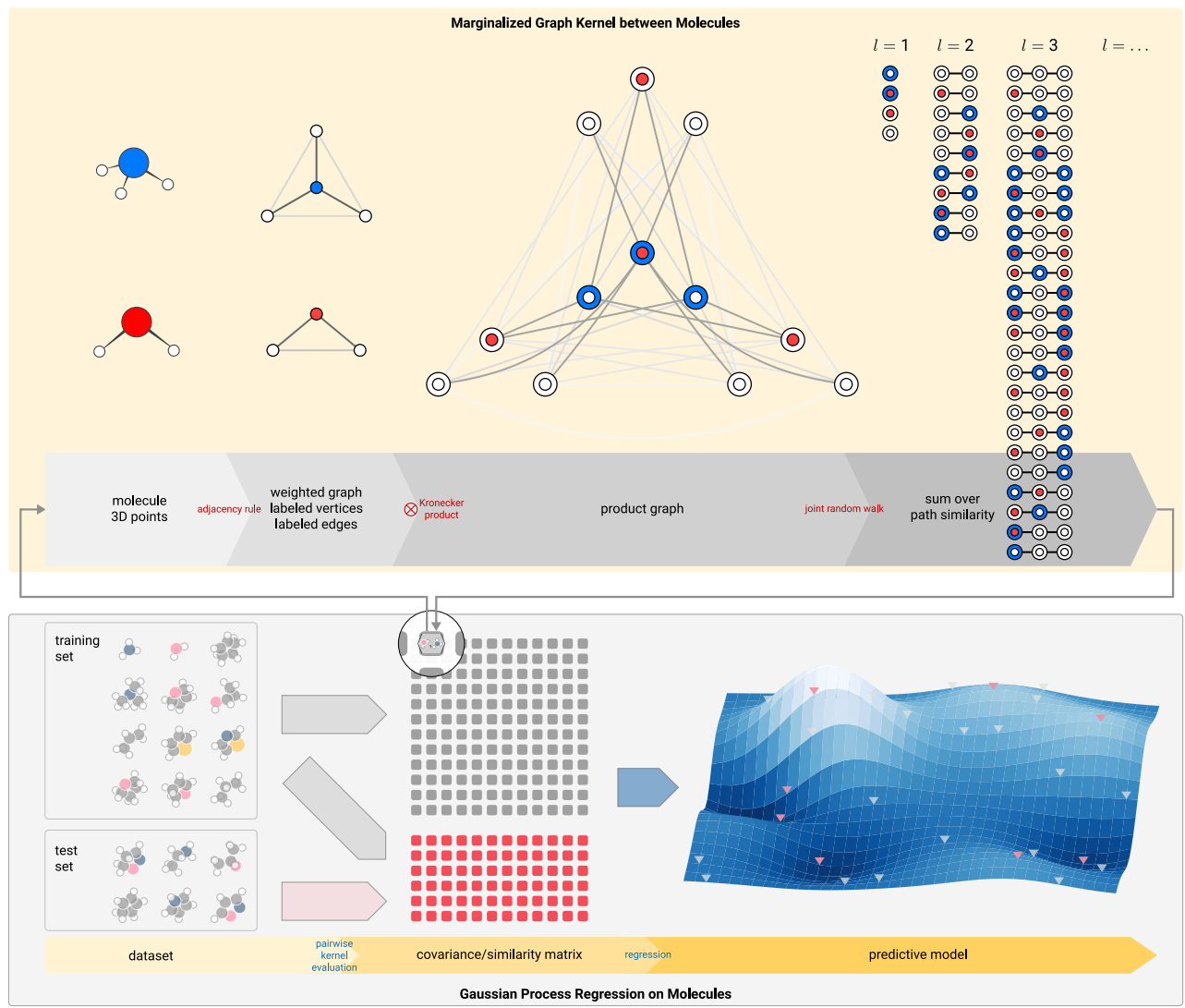

Figure 2: An overview of the proposed machine learning pipeline. **Upper:** molecules are first converted into labeled graphs with edges weighted by interatomic distances. The marginalized graph kernel defines the similarity between two graphs as the expectation of the similarity between all possible simultaneous random walk paths generated on the product graph. **Lower:** given the training molecules and unknown molecules, the pairwise similarity matrices between and within each data set are computed using the marginalized graph kernel. The training set self-similarity matrix and target values are used to construct a GPR model, which makes predictions for the unknown samples using the training-unknown cross-similarity matrix.

## 2.2 Preliminaries

We use lower case letters, *e.g.* **a**, in bold font to denote vectors, and upper case letters in bold font, *e.g.* **A**, for matrices. By default, we assume vectors are column vectors. We use $\mathbf{diag}(\mathbf{a})$ to denote a diagonal matrix whose diagonal elements are specified by **a**, and use **I** to refer to the identity matrix.

**Definition 1** *Kronecker product*

Given matrices $\mathbf{A} \in \mathbb{R}^{m_1 \times n_1}$ and $\mathbf{B} \in \mathbb{R}^{m_2 \times n_2}$, the Kronecker product $\mathbf{A} \otimes \mathbf{B} \in \mathbb{R}^{m_1 m_2 \times n_1 n_2}$ is defined as:

$$\mathbf{A} \otimes \mathbf{B} := \begin{bmatrix} \mathbf{A}_{1,1}\mathbf{B} & \mathbf{A}_{1,2}\mathbf{B} & \dots & \mathbf{A}_{1,n_1}\mathbf{B} \\ \mathbf{A}_{2,1}\mathbf{B} & \mathbf{A}_{2,2}\mathbf{B} & \dots & \mathbf{A}_{2,n_1}\mathbf{B} \\ \vdots & \vdots & \ddots & \vdots \\ \mathbf{A}_{m_1,1}\mathbf{B} & \mathbf{A}_{m_1,2}\mathbf{B} & \dots & \mathbf{A}_{m_1,n_1}\mathbf{B} \end{bmatrix}, \text{ e.g. } \begin{bmatrix} 1 & -1 \\ 2 & 10 \end{bmatrix} \otimes \begin{bmatrix} 1 & 2 \\ 3 & 4 \end{bmatrix} = \begin{bmatrix} 1 & 2 & -1 & -2 \\ 3 & 4 & -3 & -4 \\ 2 & 4 & 10 & 20 \\ 6 & 8 & 30 & 40 \end{bmatrix}$$

**Definition 2** *Hadamard (element-wise) product*

The element-wise product, also known as the Hadamard product, between two matrices of the same size $\mathbf{A}, \mathbf{B} \in \mathbb{R}^{m \times n}$ is another matrix $\mathbf{A} \odot \mathbf{B} \in \mathbb{R}^{m \times n}$ with $(\mathbf{A} \odot \mathbf{B})_{ij} := \mathbf{A}_{ij} \mathbf{B}_{ij}$.

**Definition 3** *Undirected graph*

An undirected graph $G$ is a discrete structure consisting of a set of uniquely-indexed vertices $V = \{v_1, v_2, \ldots, v_n\}$ and a set of undirected edges $E \subset V \times V$. The vertices and edges may be labeled using elements from label sets $\Sigma_v$ and $\Sigma_e$, respectively.

**Definition 4** *Weighted graph*

In a weighted graph, each edge $(v_i, v_j)$ is associated with a non-negative weight $w_{ij}$. In undirected graphs $w_{ij} = w_{ji}$. $w_{ij} = 0$ if $v_i$ and $v_j$ are not connected by an edge. An unweighted graph can be regarded as a specialized weighted graph where $w_{ij} = 1$ between each pair of $(v_i, v_j)$ connected by an edge and 0 elsewhere.

**Definition 5** *Walk on graph*

Two vertices are neighbors if they are connected by an edge. A walk on a graph is a sequence of vertices such that all pairs of adjacent vertices are neighbors.

**Definition 6** *Adjacency matrix*

The adjacency matrix of a graph of $n$ vertices is a matrix $\mathbf{A} \in \mathbb{R}^{n \times n}$ with $\mathbf{A}_{ij} = w_{ij}$. The adjacency matrices of undirected graphs are symmetric since $w_{ij} \equiv w_{ji}$.

The row-normalized adjacency matrix $\mathbf{P} := \mathbf{diag}(\mathbf{d})^{-1}\mathbf{A}$, where $\mathbf{d}_i := \sum_j \mathbf{A}_{ij}$ is the vertex degree vector, can be used as the transition matrix of a Markovian random walk process. In other words, $\mathbf{P}_{ij}$ could be interpreted as the conditional probability for a random walker at vertex $v_i$ to jump to $v_j$ during a single step.

## 2.3 Graph Kernel for Molecules

### 2.3.1 Graph Representations for Molecules

The practice of using labeled graphs, with the exemplary ball-and-stick model, to represent molecules has gained popularity well before the era of machine learning. In this work, we represent a molecule of $n$ atoms as an undirected graph $G = \left\{ V = \{v_i\}, E = \{e_{ij}\} \right\}$, $i, j \in \{1 \ldots n\}$, where atoms are represented by vertices $v_i \in \{H, C, N, O, \ldots\}$ that are labeled by chemical elements. Each edge $e_{ij} \in \mathbb{R}$ between vertices $i$ and $j$ is labeled by the distance between the atoms, while its weight $w_{ij}$ is set by a spatial adjacency rule $\mathcal{A}(\mathbf{r}_i, \mathbf{r}_j)$, whose specific form will be discussed in section 3.1. The adjacency matrix of a molecular graph is thus $\mathbf{A}_{ij} = \mathcal{A}(\mathbf{r}_i, \mathbf{r}_j)$. Note that the edges are often supersets of the collection of covalent bonds in a molecule.

### 2.3.2 Marginalized Graph Kernel

The marginalized graph kernel $K(G, G')$ [24] defines the overall similarity between two graphs as the expectation of the path similarity between all pairs of paths that can be obtained by performing simultaneous random walks on two graphs $G$ and $G'$. The explicit formula for the path similarity expectation takes the form:

$$K(G, G') = \sum_{\ell=1}^{\infty} \sum_{\mathbf{h}} \sum_{\mathbf{h}'} \Bigg[ p_s(h_1)\, p'_s(h'_1)\, K_v(v_{h_1}, v'_{h'_1})\, p_q(h_\ell)\, p'_q(h'_\ell) \\
\times \left( \prod_{i=2}^{\ell} p_t(h_i | h_{i-1}) \right) \\
\times \left( \prod_{j=2}^{\ell} p'_t(h'_j | h'_{j-1}) \right) \\
\times \left( \prod_{k=2}^{\ell} K_v(v_{h_k}, v'_{h'_k}) K_e(e_{h_{k-1} h_k}, e'_{h'_{k-1} h'_k}) \right) \Bigg]. \tag{1}$$

Here, $\ell$ is the length of the path, $\mathbf{h}$ and $\mathbf{h}'$ are paths on the graphs represented by length-$\ell$ vectors of vertex labels, $p_s(\cdot)$ is the starting probability of the random walk on each vertex, $p_q(\cdot)$ is the stopping probability of the random walk on each vertex at any given step, $p_t(\cdot, \cdot)$ is the transition probability between a pair of vertices, $K_v(\cdot, \cdot)$ is an elementary kernel that computes the similarity between two vertices, $K_e(\cdot, \cdot)$ is another elementary kernel that computes the similarity between pairs of bonds.

Despite the overwhelming appearance of the infinite summation, eq. (1) can be reformulated under the spirit of dynamic programming as:

$$K(G, G') = \sum_{h_1 \in V, h'_1 \in V'} p_s(h_1)\, p'_s(h'_1)\, K_v(h_1, h'_1)\, R_\infty(h_1, h'_1), \tag{2}$$

where $R_\infty$ is the solution to the linear system:

$$R_\infty(h_1, h'_1) = p_q(h_1)\, p'_q(h'_1) + \sum_{i \in V, j \in V'} t(i, j, h_1, h'_1)\, R_\infty(i, j), \tag{3}$$

with

$$t(i, j, h_1, h'_1) := p_t(i|h_1)\, p'_t(j|h'_1)\, K_v(v_i, v_j)\, K_e(e_{i\, h_1}, e_{j\, h'_1}). \tag{4}$$

Equations (2) to (4) in fact exhibit a Kronecker product structure, which can be readily recognized in matrix

form:

$$K(G, G') = \left(\mathbf{p} \otimes \mathbf{p}'\right)^\mathsf{T} \cdot \mathbf{diag}\left(\mathbf{v} \overset{K_v}{\otimes} \mathbf{v}'\right) \cdot \mathbf{r}_\infty, \tag{5}$$

with $\mathbf{r}_\infty$ being the solution to the linear system

$$\mathbf{r}_\infty = \mathbf{q} \otimes \mathbf{q}' + \left[\left(\mathbf{P} \otimes \mathbf{P}'\right) \odot \left(\mathbf{E} \overset{K_e}{\otimes} \mathbf{E}'\right)\right] \cdot \mathbf{diag}\left(\mathbf{v} \overset{K_v}{\otimes} \mathbf{v}'\right) \cdot \mathbf{r}_\infty, \tag{6}$$

where

- $\mathbf{v}$ is the vertex label vector of $G$ with $\mathbf{v}_i = v_i$;
- $\mathbf{p}$ is the starting probability vector of $G$ with $\mathbf{p}_i = p_s(v_i)$;
- $\mathbf{q}$ is the stopping probability vector of $G$ with $\mathbf{q}_i = p_q(v_i)$;
- $\mathbf{P}$ is the transition probability matrix as defined in definition 6, where the augmented degree $\tilde{\mathbf{d}}_i = \sum_j \mathbf{A}_{ij}/(1-\mathbf{q}_i)$ ensure that the transition and stopping probabilities on each node sum to 1;
- $\mathbf{E}$ is the edge label matrix of $G$ with $\mathbf{E}_{ij} = e_{ij}$;
- $\mathbf{v}', \mathbf{p}', \mathbf{q}', \mathbf{P}', \mathbf{E}'$ are the corresponding vectors and matrices for $G'$;
- $\overset{K_v}{\otimes}$ is the kernelized Kronecker product between $\mathbf{v}$ and $\mathbf{v}'$ where $K_v(\cdot, \cdot)$ replaces the product operation between the vertex labels;
- $\overset{K_e}{\otimes}$ is the kernelized Kronecker product between $\mathbf{E}$ and $\mathbf{E}'$ where $K_e(\cdot, \cdot)$ replaces the product operation between the edge labels.

For clarity of the discussion below, we denote

$$\mathbf{V}_\times := \mathbf{diag}\left(\mathbf{v} \overset{K_v}{\otimes} \mathbf{v}'\right),$$
$$\mathbf{D}_\times := \mathbf{diag}(\tilde{\mathbf{d}}) \otimes \mathbf{diag}(\tilde{\mathbf{d}}'),$$
$$\mathbf{A}_\times := \mathbf{A} \otimes \mathbf{A}',$$
$$\mathbf{P}_\times := \mathbf{P} \otimes \mathbf{P}' = \mathbf{D}_\times^{-1} \mathbf{A}_\times,$$
$$\mathbf{E}_\times := \mathbf{E} \overset{K_e}{\otimes} \mathbf{E}',$$
$$\mathbf{p}_\times := \mathbf{p} \otimes \mathbf{p}',$$
$$\mathbf{q}_\times := \mathbf{q} \otimes \mathbf{q}'.$$

To solve eq. (6), first observe that only the product $\mathbf{V}_\times \mathbf{r}_\infty$ as a whole is needed to compute $K(G, G')$. We can thus rearrange eq. (6) to form a symmetric linear system.

$$\mathbf{r}_\infty - (\mathbf{P}_\times \odot \mathbf{E}_\times) \mathbf{V}_\times \mathbf{r}_\infty = \mathbf{q}_\times \tag{7}$$

$$\left(\mathbf{V}_\times^{-1} - \mathbf{P}_\times \odot \mathbf{E}_\times\right) \mathbf{V}_\times \mathbf{r}_\infty = \mathbf{q}_\times \tag{8}$$

$$\mathbf{V}_\times \mathbf{r}_\infty = \left(\mathbf{V}_\times^{-1} - \mathbf{P}_\times \odot \mathbf{E}_\times\right)^{-1} \mathbf{q}_\times \tag{9}$$

$$= \left[\mathbf{V}_\times^{-1} - \left(\mathbf{D}_\times^{-1} \mathbf{A}_\times\right) \odot \mathbf{E}_\times\right]^{-1} \mathbf{q}_\times \tag{10}$$

$$= \left(\mathbf{D}_\times \mathbf{V}_\times^{-1} - \mathbf{A}_\times \odot \mathbf{E}_\times\right)^{-1} \mathbf{D}_\times \mathbf{q}_\times \tag{11}$$

The linear system $\mathbf{D}_\times \mathbf{V}_\times^{-1} - \mathbf{A}_\times \odot \mathbf{E}_\times$ in eq. (11) is symmetric and positive-definite, as long as $q > 0$, $K_v(\cdot, \cdot) <= 1$, and $K_e(\cdot, \cdot) <= 1$. It can be solved efficiently using an iterative method such as conjugate gradient [25, 26]. The full expression for the marginalized graph kernel in matrix form is then

$$K(G, G') = \mathbf{p}_\times^\mathsf{T} \left( \mathbf{D}_\times \mathbf{V}_\times^{-1} - \mathbf{A}_\times \odot \mathbf{E}_\times \right)^{-1} \mathbf{D}_\times \mathbf{q}_\times \tag{12}$$

'Stray' atoms, *i.e.* atoms that are not neighbors to any other atoms, have $\mathbf{d}_i = 0$. This pathological situation can render the linear system singular, and is resolved by setting the corresponding $\mathbf{d}_i = \mathbf{q}_i = 1$. This is equivalent to immediately terminating any random walk path originating from the stray atoms.

## 2.4 Gaussian Process Regression and Active Learning

### 2.4.1 Energy Regression

Given a training set $D$ of $m$ molecules and their associated energy $\{(M_1, \ldots, M_m), (E_1, \ldots, E_m)\}$, as well as a marginalized graph kernel $K$, the GPR prediction for the energy $\{E_1^*, \ldots, E_n^*\}$ of a test set of $n$ unknown molecules $\{M_1^*, \ldots, M_n^*\}$ can be derived analytically as

$$\mathbf{E}^* := [E_1^*, \ldots, E_n^*]^\mathsf{T} = \mathbf{K}_{D*}^\mathsf{T} \mathbf{K}_{DD}^{-1} \mathbf{y}_D, \tag{13}$$

which is identical to that in kernel ridge regression [27]. The predictive uncertainty is given by the posterior covariance matrix

$$\mathbf{\Sigma}^* = \mathbf{K}_{**} - \mathbf{K}_{D*}^\mathsf{T} \mathbf{K}_{DD}^{-1} \mathbf{K}_{D*}. \tag{14}$$

Here,

$$\mathbf{K}_{DD}^{n \times n}(i, j) = K(M_i, M_j), \tag{15}$$

$$\mathbf{K}_{D*}^{n \times m}(i, j) = K(M_i, M_j^*), \tag{16}$$

$$\mathbf{K}_{**}^{m \times m}(i, j) = K(M_i^*, M_j^*) \tag{17}$$

are the training set covariance matrix, training-test cross-covariance matrix, and test set covariance matrix, respectively, that are obtained by iterating the graph kernel over all pairs of molecules from the corresponding datasets.

The marginalized graph kernel is particularly suitable for predicting extensive properties such as molecular energy. To see why, first recall that the kernel possesses a vertex-wise summation structure $\sum_{h, h'} \cdots$ as manifested in eq. (2), and that $R_\infty(h_1, h_1')$ can be alternatively expressed as

$$R_\infty(h_1, h_1') = \sum_{\ell=1}^{\infty} r_\ell(h_1, h_1') \tag{18}$$

with

$$r_1(h_1, h'_1) = p_q(h_1)\, p'_q(h'_1), \tag{19}$$

$$r_\ell(h_1, h'_1) = \sum_{h_2, h'_2} \left[ t(h_2, h'_2, h_1, h'_1) \sum_{h_3, h'_3} \left[ t(h_3, h'_3, h_2, h'_2) \sum_{\cdots} \left[ \cdots \sum_{h_\ell, h'_\ell} t(h_\ell, h'_\ell, h_{\ell-1}, h'_{\ell-1})\, p_q(h_\ell)\, p'_q(h'_\ell) \right] \right] \right]. \tag{20}$$

Equation (2), eq. (18), and eq. (20) together indicate that each of the $p_s(h_1)\, p'_s(h'_1)\, K_v(h_1, h'_1)\, R_\infty(h_1, h'_1)$ terms effectively computes the expectation of the similarity between all random walk paths originating from the pair of vertices $(h_1, h'_1)$. We could thus regard it as a vertex-wise atomistic neighborhood similarity kernel, denoted as $\kappa(\cdot, \cdot)$:

$$\kappa(h_1, h'_1) \doteq p_s(h_1)\, p'_s(h'_1)\, K_v(h_1, h'_1)\, R_\infty(h_1, h'_1). \tag{21}$$

The overall similarity between two graphs as computed by the marginalized graph kernel is thus merely a summation over the similarity between all pairs of vertices, *i.e.*

$$K(G, G') = \sum_{v, v'} \kappa(v, v'). \tag{22}$$

From this perspective, the graph kernel can also be viewed as a global structural kernel, as mentioned in [28], that sums over the environment covariance matrix computed between atomistic neighborhoods.

To ensure extensiveness of the energy prediction, we can hypothetically predict the individual contributions of each atom to the total energy of an unknown molecule, and then sum up the contributions. We assume that there exists some energy decomposition algorithm (EDA) [29] that could localize the energy of a molecule among its atoms as $\{e_1, e_2, \ldots\} = \mathbf{EDA}(E; M)$ subject to $\sum_i e_i = E$. Note that the EDA scheme is purely formal and is never explicitly used, as will be shown shortly. Using $\kappa(\cdot, \cdot)$ as a covariance function, we obtain the following GPR formulation:

$$\begin{bmatrix} e^*_1 \\ e^*_2 \\ \vdots \\ e^*_n \end{bmatrix} = \begin{bmatrix} \kappa(v^*_1, v^1_1) & \kappa(v^*_2, v^1_1) & \cdots \\ \kappa(v^*_1, v^1_2) & \kappa(v^*_2, v^1_2) & \cdots \\ \vdots & \vdots & \vdots \\ \kappa(v^*_1, v^2_1) & \kappa(v^*_2, v^2_1) & \cdots \\ \kappa(v^*_1, v^2_2) & \kappa(v^*_2, v^2_2) & \cdots \\ \vdots & \vdots & \vdots \end{bmatrix}^\mathsf{T} \begin{bmatrix} \kappa(v^1_1, v^1_1) & \kappa(v^1_1, v^1_2) & \cdots & \kappa(v^1_1, v^2_1) & \kappa(v^1_1, v^2_2) & \cdots & \cdots \\ \kappa(v^1_2, v^1_1) & \kappa(v^1_2, v^1_2) & \cdots & \kappa(v^1_2, v^2_1) & \kappa(v^1_2, v^2_2) & \cdots & \cdots \\ \vdots & \vdots & \ddots & \vdots & \vdots & \cdots & \cdots \\ \kappa(v^2_1, v^1_1) & \kappa(v^2_1, v^1_2) & \cdots & \kappa(v^2_1, v^2_1) & \kappa(v^2_1, v^2_2) & \cdots & \cdots \\ \kappa(v^2_2, v^1_1) & \kappa(v^2_2, v^1_2) & \cdots & \kappa(v^2_2, v^2_1) & \kappa(v^2_2, v^2_2) & \cdots & \cdots \\ \vdots & \vdots & \vdots & \vdots & \vdots & \ddots & \cdots \\ \vdots & \vdots & \vdots & \vdots & \vdots & \vdots & \ddots \end{bmatrix}^{-1} \begin{bmatrix} e^1_1 \\ e^1_2 \\ \vdots \\ e^2_1 \\ e^2_2 \\ \vdots \\ \vdots \end{bmatrix}, \tag{23}$$

where the superscripts correspond to molecule ids.

Under the multivariate normality condition, the total energy $E$, being the sum of individual random variables representing atomic energies, is itself a normally distributed random variable, The variance of $E$ is $\text{var}[E] = \text{var}[\sum_i e_i] = \sum_{ij} \text{cov}(e_i, e_j)$, while the covariance between $E$ and $E'$ is $\text{cov}[E, E'] = \text{cov}[\sum_i e_i, \sum_j e'_j] = \sum_{ij} \text{cov}[e_i, e'_j]$. Note that this exhibits exactly the same pairwise summation structure as that of the marginalized graph kernel. Thus, by summing up the cross-covariance submatrices between pairs of molecules in eq. (23) and substituting $K(\cdot, \cdot)$ for $\sum_{ij} \kappa(v_i, v'_j)$, we could define an alternative GPR model that bypasses energy decomposition:

$$E^* = \begin{bmatrix} K[M^*, M_1] \\ K[M^*, M_2] \\ \vdots \end{bmatrix}^\mathsf{T} \begin{bmatrix} K[M_1, M_1] & K[M_1, M_2] & \ldots \\ K[M_2, M_1] & K[M_2, M_2] & \ldots \\ \vdots & \vdots & \ddots \end{bmatrix}^{-1} \begin{bmatrix} E_1 \\ E_2 \\ \vdots \end{bmatrix}. \tag{24}$$

This leads exactly back to eq. (13). Thus, we have shown that when only the total energy is of interest, a GPR prediction for the energy of an entire molecule using the marginalized graph kernel is equivalent to a summation of the GPR predictions for the localized atomic contributions using an atomistic neighborhood kernel. The former approach, however, avoids the potentially costly and controversial explicit energy decomposition, while still enjoying automatic scaling of the predicted energy with respect to molecule size. This approach can also speed up GPR computation by creating covariance matrices whose size is proportional to the number of molecules, rather than the number of total atoms, in the data set.

### 2.4.2 Active Learning Protocol

Given a training set $S$, a test set $T$, an acquisition function $\mathcal{Q}(\cdot, \cdot)$ that measures the learning value of a sample given a GPR model, active learning can be performed using the following protocol:

1: **repeat**
2:     $G \leftarrow \text{TrainGPR}(S)$,
3:     $\alpha \leftarrow \underset{i \in T}{\text{argmax}}\ \mathcal{Q}(G, i)$,
4:     $S \leftarrow S \cup \{\alpha\}$,
5:     $T \leftarrow T \setminus \{\alpha\}$,
6: **until** stop criteria met.

A natural choice of the acquisition function would be prediction error, which is the absolute difference between the GPR prediction and the truth value of the target function at the sample points. However, since the ground truth may not be known or is too difficult to obtain for every point in practice, the GPR predictive uncertainty could be used as an alternative acquisition function. In this case, the 'unsupervised' learning procedure can proceed even without any target function value due to the fact that the GPR posterior variance does not depend on the target value.

# 3 Computation and Results

## 3.1 Kernel Specification and Data Set

The vertex elementary kernel that we use for the marginalized graph kernel is an elevated Kronecker delta function on element symbols:

$$K_v(v, v') = \begin{cases} 1, & \text{if } v = v', \\ \nu \in (0, 1), & \text{otherwise}. \end{cases} \quad (25)$$

The edge elementary kernel is a square exponential function on edge lengths, which evaluates to 1 if two edges are of the same length and smoothly transitions to 0 as the difference in lengths grows:

$$K_e(e, e') = \exp\left[-\frac{1}{2}\frac{(e-e')^2}{\lambda^2}\right]. \quad (26)$$

The adjacency rule that computes the weights for each edge also assumes a square exponential form:

$$\mathcal{A}(\mathbf{r}_i, \mathbf{r}_j) = \exp\left[-\frac{1}{2}\frac{\|\mathbf{r}_i - \mathbf{r}_j\|^2}{(\zeta\,\sigma_{ij})^2}\right], \quad (27)$$

where $\sigma_{ij}$, as given in the appendix, are element-wise length scale parameters derived from typical bonding lengths. A uniform starting probability $p_s(\cdot) \equiv s$ and a uniform stopping probability $p_q(\cdot) \equiv q$ are used across all vertices.

In fig. 3, we visualize in matrix form the atomistic neighborhood similarities $\mathbf{V}_\times \mathbf{r}_\infty$ between a methoxyethane ($CH_3CH_2OCH_3$) molecule and a 2-ethoxyethanol ($CH_3CH_2OCH_2CH_2OH$) molecule, computed with $\nu = 0.25$, $\zeta = 1$, $\lambda = 0.02$, $s = 1$, $q = 0.01$. Note that the marginalized graph kernel can, for example, differentiate between atoms of the same element yet are embedded in different local chemical environments.

The QM7 dataset, which contains 7165 molecules with up to 7 heavy atoms and 23 total atoms along with their atomization energies, is used as the data set for benchmarking purposes [11, 30].

## 3.2 Hyperparameter Selection

The complete marginalized graph kernel as specified in section 3.1 is controlled by 5 hyperparameters $\nu$, $\zeta$, $\lambda$, $s$, and $q$, each of which possesses a unique physical interpretation:

- $\nu$ is the baseline, or 'prior', similarity that we assign between atoms of difference elements. $\nu$ must be within $(0, 1)$ to ensure that the linear system in eq. (11) is positive-definite.
- $\zeta$ controls how quickly the weight of an edge decays with regard to increasing interatomic distance. It affects the edge density of the molecular graph and is one of the two factors that determines the neighborhood range of each atom.
- $\lambda$ control the sensitivity of the kernel with respect to differences in edge length. Since the order of a covalent bond is strongly correlated with its length, $\lambda$ can be effectively used to discriminate different

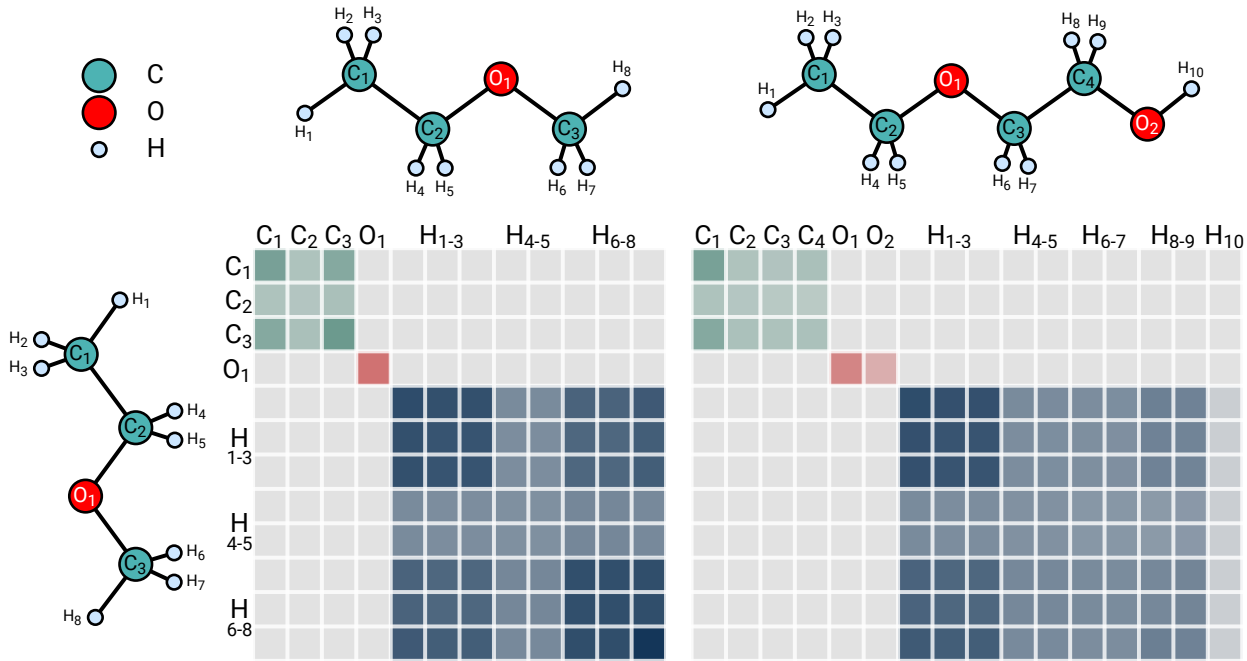

Figure 3: An illustration of the marginalized graph kernel computation result among two molecules: methoxyethane $CH_3CH_2OCH_3$ and 2-ethoxyethanol $CH_3CH_2OCH_2CH_2OH$. Each square tile corresponds to an element of the $\mathbf{V}_\times \mathbf{R}_\infty$ vector, which is a weighted average (marginalization) of the similarity between all paths generated by simultaneous random walks originating from the two atoms. The matrices effectively visualizes the atomistic neighborhood similarities with darker colors indicate higher similarity. Note for example the difference in similarity between the oxygen atoms in the ether bond to itself and to the oxygen atom in the alcohol group.

types of bonds.

- $s$ affects the sampling preference of the simultaneous random walk among pairs of vertices, and could thus be thought as a factor of significance. From the perspective of GPR modeling, the magnitude of $s$ is the *a priori* standard deviation of atomic contributions to molecular energy. In practice, we may abandon the probabilistic interpretation of $s$ and allow it to assume values that do not sum up to unity.
- $q$ determines the average length of the random walk paths being sampled, and together with $\zeta$ determines the range of the atomistic neighborhoods that are effectively compared between atoms.

Maximum likelihood estimation (MLE) is commonly regarded as the 'default' method for selecting hyperparameters in Bayesian inference. However, this should be exercised with caution because certain choices of the hyperparameters could arbitrarily maximize the log-likelihood function without actually improving the predictive power of the GPR model. To obtain an insight into the influence of the hyperparameters on model performance, we performed an exhaustive scan of the hyperparameters within the range $\nu \in [0.1, 0.9], \zeta \in [0.5, 1.5], \lambda \in [0.01, 0.4], s \in [10, 500], q \in [0.01, 0.5]$. The mean absolute error (MAE)-likelihood joint distributions, as shown in the upper panels of fig. 4, reveal the fact that likelihood alone does not necessarily indicate high prediction accuracy. This is manifested by the noticeable amount of hyperparameter sets that lies on the upper left parts of the joint distribution density. Aside from that, as shown by the lower panels of fig. 4, there are fortunately many hyperparameter sets that exists in the upper left parts of the joint distributions between MAE and error-uncertainty correlation. These hyperparameters lead to GPR models with both high predictive accuracy and uncertainty-error correlation, the latter of which is crucial for unsupervised active learning.

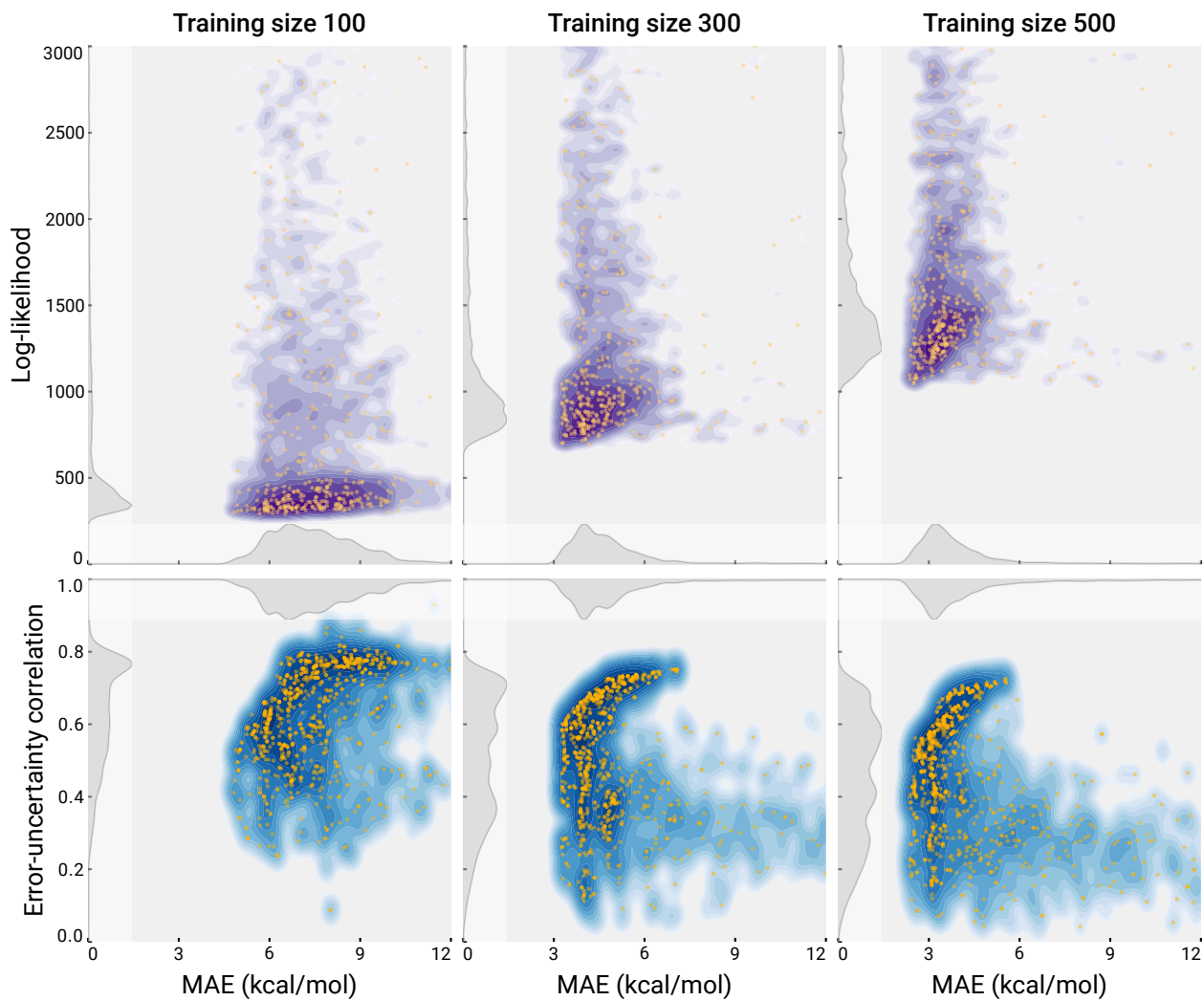

Figure 4: **Upper:** The extended vertical tail of the error-likelihood joint distribution indicate that GPR models with high log-likelihood could still possess good predictive power. **Lower:** Despite that the correlation between the predictive uncertainty and error generally decrease as the accuracy of the GPR models improve, it is still always possible to find sets of hyperparameters that lie close to the upper left corner of the plots, which are the regions corresponding to relatively low errors and high uncertainty-error correlation. The latter property is crucial for effective unsupervised active learning. The densities are estimated using a grid of 5400 hyperparameters sets.

A further examination on the conditional distribution of the mean absolute error, log-log-likelihood, and error-uncertainty correlation provides additional insights into the effect of the hyperparameters. As shown by fig. 5, the edge kernel length scale parameter $\lambda$ appears to be the single most sensitive hyperparameter, followed by $\zeta$, $\nu$, and $q$, that affects both predictive accuracy and error-uncertainty correlation. A uniform starting probability $s$ seems to only affect the model likelihood. However, $s$ actually scales the magnitude of the predictive uncertainty, which in turn controls the width of the GPR predictive confidence interval. As shown in fig. 6, we examine how $s$ affects the quality of the predictive uncertainty as an alternative to the true error, which may not be known in practice, by checking if the predictive confidence intervals contain a proportional percentage of the true value of the test samples. For example, an interval surrounding the predictive mean that corresponds to a 50% confidence level should roughly contain the true value of 50% of the test samples. Overall, we found that a proper choice of $s$ can lead to a coverage curve that lies very close to the diagonal. Meanwhile, the GPR model has a general tendency to be overfident with large training sets, most likely due to the existence of molecules that consistently cause large errors as demonstrated later in the lower panel of fig. 7. In the subsequent section, we use $\zeta = 1$, $\nu = 0.3$, $\lambda = 0.05$, $s = 250$, and $q = 0.05$ for all GPR modeling tasks.

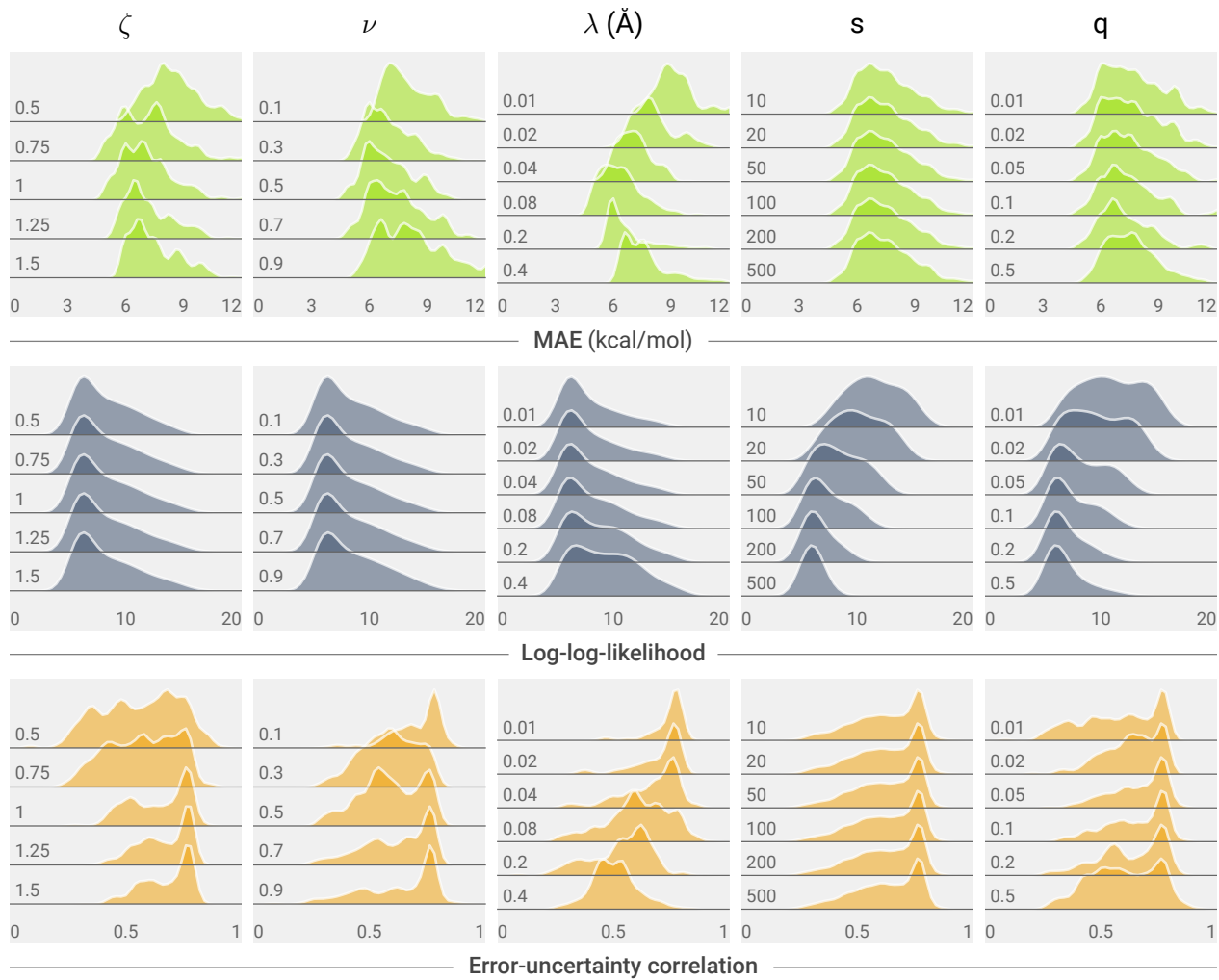

Figure 5: Shown here is the distribution of the mean absolute error, log-log-likelihood, and error-uncertainty correlation of GPR models conditioned on each specific value of the hyperparameters. 300 molecules are randomly selected from the QM7 dataset as the training set, while the remaining 6865 molecules form the test set. The accuracy and error-uncertainty correlation of the resulting GPR model are influenced mostly by $\zeta$, $\nu$, $\lambda$, and $q$, while the log-likelihood function is most strongly affected by $s$ and $q$. The choice of the optimal hyperparameter apparently becomes a subtle problem due to the non-trial dependency between accuracy and error-uncertainty correlation.

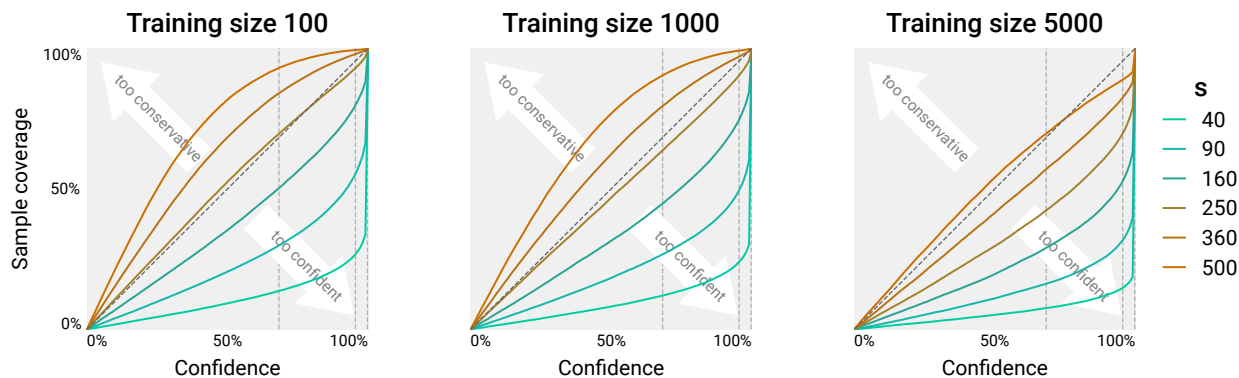

Figure 6: The generalized starting probability parameter $s$ determines the width of the predictive confidence interval, which should contain a proportional percentage of the ground truth energy of the test samples. Smaller $s$ values lead to aggressive confidence interval estimations, yielding coverage curves below the diagonal. Larger $s$ values, in contrast, lead to conservative estimations and wide confidence intervals, yielding curves that lie above the diagonal.

## 3.3 Energy Prediction

As a first benchmark of prediction accuracy, we use a random selection of $N$ molecules from the QM7 dataset to train GPR models and measure the MAE and root-mean square error (RMSE) on energy prediction for the rest of the molecules. As shown in the upper left panel of fig. 7 and in Table 1, using 15 randomly selected training sets, the presented method delivers an MAE of $1.01 \pm 0.04$ kcal/mol when $N = 5000$. This is comparable to the best of existing methods [13, 28, 31–35]. It also achieves an MAE of $1.48 \pm 0.05$ kcal/mol on a significantly smaller $N = 2000$. The training time ranges between $\sim 0.3$ s for $N = 100$ to $\sim 380$ s for $N = 5000$ on an in-house GPU-accelerated implementation, while the prediction time ranges between 0.005 s/sample with 100 training samples to 0.086 s/sample with 5000 training samples.

We further examined the performance of the method for carrying out the active learning protocol as described in section 2.4.2. In each parallel test, the process starts with a seeding set of 50 training samples and iteratively adds the test sample with the highest acquisition score into the training sets until the $N$ reaches 2000. As shown in the upper right panel of fig. 7 and in Table 1, supervised active learning, which uses the true predictive error as the acquisition function, leads to GPR models with significantly lower MAE and RMSE than those trained on a randomly selected training set. Averaging over 7 parallel runs, we obtain MAE and RMSE of $0.62 \pm 0.01$ kcal/mol and $0.76 \pm 0.01$ kcal/mol, respectively. Meanwhile, unsupervised active learning, which uses the predictive uncertainty as the acquisition function, seems to perform only slightly better than those trained on random training sets with an MAE of $1.28 \pm 0.03$ kcal/mol, but does achieve a significantly lower RMSE of $1.81 \pm 0.06$ kcal/mol.

In the lower panel of fig. 7, we visualize some of the molecules that consistently cause large prediction errors even with the various GPR models trained with up to 5000 molecules. It is easy to recognize that a common feature present in molecule A-D is the three-membered ring, which is a highly distorted structure. Molecule E and F contain a conjugated triple bond system and an unsaturated bicyclic ring with an oxygen bridge, respectively, which are rare in the benchmark dataset. The lack of explicit angular information in

Table 1: Energy prediction accuracy of the present and previously published methods.

| Training set | Representation | Kernel | Regression | MAE (kcal/mol) | RMSE (kcal/mol) | Source |
| --- | --- | --- | --- | --- | --- | --- |
| 2000, random | molecular graph | graph kernel | GPR | 1.48±0.05 | 3.57±0.46 | present |
| 2000, supervised | molecular graph | graph kernel | GPR | 0.62±0.01 | 0.76±0.01 | present |
| 2000, unsupervised | molecular graph | graph kernel | GPR | 1.28±0.03 | 1.81±0.06 | present |
| 2000 | CM | Laplacian | KRR | 4.32 | - | [31] |
| 5000, random | molecular graph | graph kernel | GPR | 1.01±0.04 | 2.29±0.57 | present |
| 5000 | MBTR | Gaussian | KRR | 0.60 | 0.97 | [33] |
| 5000 | SOAP | REMatch | KRR | 0.92 | 1.61 | [28] |
| 5000 | BAML | Laplacian | KRR | 1.15 | 2.54 | [36] |
| 5732 | CM | Laplacian | KRR | 3.07±0.07 | 4.84±0.40 | [31] |
| 5732 | BoB | Laplacian | KRR | 1.50 | - | [32] |
| 5732 | BoB | Gaussian | KRR | 2.40 | - | [13] |
| 5732 | CM | Laplacian | KRR | 3.37 | - | [13] |
| 5732 | EB | Gaussian | KRR | 1.19 | - | [13] |
| 5768 | nuclear charge & interatomic distance | - | DTNN | 1.04 | 1.43 | [34] |

CM: Coulomb Matrix; MBTR: many-body tensor representation;
SOAP: smooth overlap of atomic positions; BoB: bag of bonds; EB: encoded bond.

our current implementation could have weakened the performance of the graph kernel in such situations. Nonetheless, the GPR model is potentially able to use high predictive uncertainties as a form of alert to trigger fall-back mechanisms such as alternative calculations.

## 4 Conclusion

In this paper, we presented a new machine learning pipeline, which integrates the marginalized graph kernel, the Gaussian process regression method, and an active learning protocol, for predicting molecular atomization energies. The method achieves excellent accuracy while using significantly smaller training sets as compared with previous methods. We show that the marginalized graph kernel defines a molecular similarity metric that is computed using both topological and geometric information, and that the kernel can naturally adapt to molecules containing diverse topology, number of atoms, and element species. GPR models created using the marginalized graph kernel can produce energy predictions that scale properly with molecule size while bypassing any explicit energy decomposition procedure. We demonstrate that the convolutional — or pairwise summation — structure of the kernel is crucial in enabling the prediction of extensive properties that automatically scales with molecule size. It is worth noting that the work is inspired by the localized random walk graph kernel for molecules proposed by Ferré *et al* [35], which is essentially a localized marginalized graph kernel equipped with constant vertex and edge kernels.

We expect that the presented method could benefit from a more thorough design and selection of the hyperparameters involving, for example, pairwise priors that are tailored between each type of atoms and bonds. Incorporation of explicit angular information into the vertices should greatly improve the ability of the kernel to recognize and differentiate molecules with highly strained geometry. A wider adoption of the graph kernel entail a more efficient computer implementations that can work around the quartic asymptotic computational cost for solving the Kronecker product system induced by large molecules such as polymers, proteins, and polynucleotides.

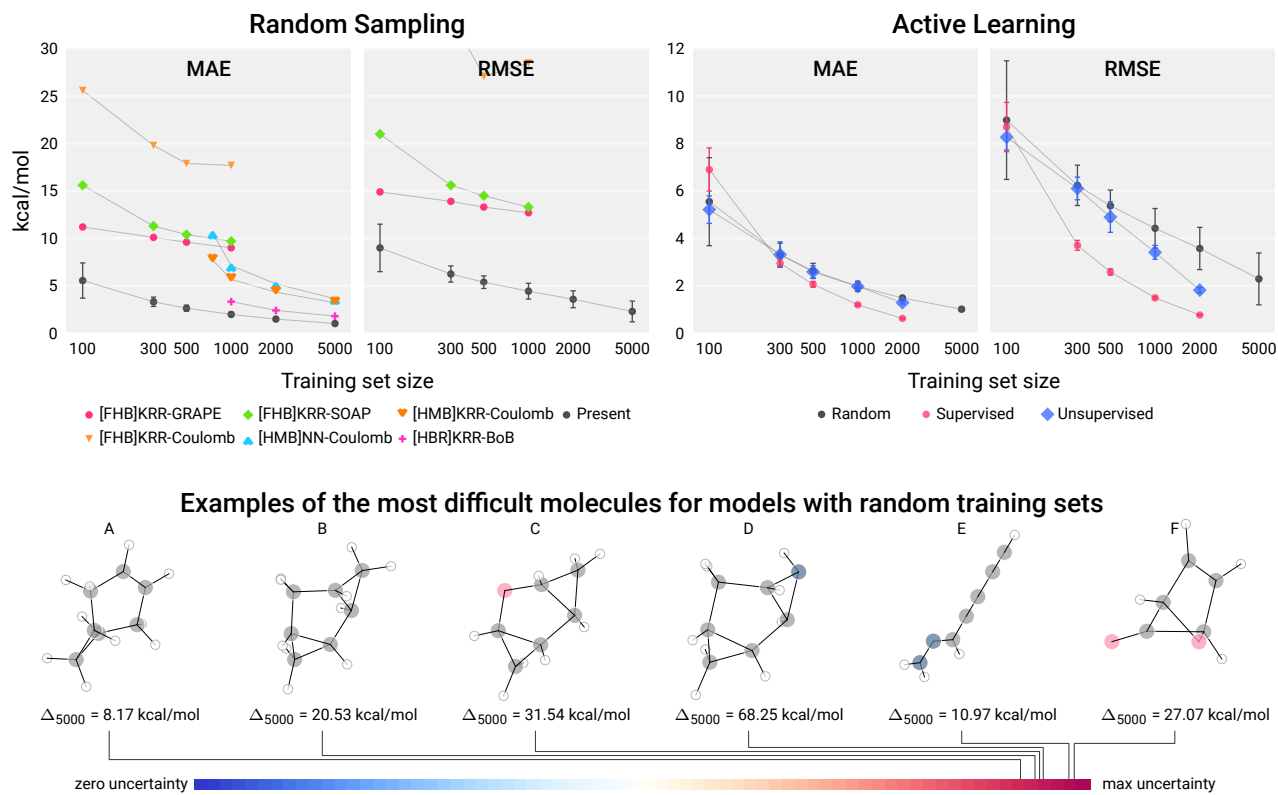

Figure 7: **Upper left:** Saturation curves for the mean absolute error and root-mean square error of the presented method and previous methods on predicting the atomization energy of molecules in the QM7 dataset. The data points for the graph kernel are averaged over 15 random training sets and 7 active learning realizations, respectively. Vertical bars indicate standard deviation. Reference designation: FHB [35], HMB [31], HBR [32]. **Upper right:** Comparison of the performance between GPR models built from random training sets and those built from an active learning process. **Lower:** Even with models trained on 5000 randomly chosen samples, some molecules can consistently cause large prediction error. This behavior is attributed to highly strained geometric features such as 3-rings and conjugated triple bonds. Fortunately, the GPR model is able to use high predictive uncertainties to inform users of such situations , and to prioritize the inclusion of those molecules during active learning.

# Acknowledgment

This work was supported by the Luis W. Alvarez Postdoctoral Fellowship at Lawrence Berkeley National Laboratory. YHT appreciates advice from Timur Takhtaganov regarding predictive confidence interval analysis.

# Appendix

### Length scale parameters in the square exponential adjacency rule

The length scale parameter $\sigma_{ij}$, which are used in the square exponential adjacency rule to weight the edges, are given in table 2.

Table 2: Common bond lengths averaged over multiple sources [37–40].

|   | H | C | O | N | F | S |
|---|---|---|---|---|---|---|
| H | 0.74 | 1.09 | 0.96 | 1.01 | 0.92 | 1.34 |
| C | 1.09 | 1.39 | 1.27 | 1.34 | 1.35 | 1.82 |
| O | 0.96 | 1.27 | 1.48 | 1.23 | 1.42 | 1.44 |
| N | 1.01 | 1.34 | 1.23 | 1.26 | 1.38 | 1.68 |
| F | 0.92 | 1.35 | 1.42 | 1.38 | 1.42 | 1.57 |
| S | 1.34 | 1.82 | 1.44 | 1.68 | 1.57 | 2.05 |